\documentclass[sigconf, nonacm, pdfa=true]{acmart}
\renewcommand\footnotetextcopyrightpermission[1]{}

\usepackage{tikz}
\usepackage{pgfplots}
\usepackage{caption}
\pgfplotsset{compat=1.18}

\usepackage{microtype}
\usepackage{siunitx}

\begin{document}

\title{Prosthetic Hand Manipulation System Based on EMG and Eye Tracking Powered by the Neuromorphic Processor AltAi}

\author{Roman Akinshin}
\orcid{0009-0006-2632-9922}
\affiliation{%
  \institution{Skolkovo Institute of Science and Technology}
  \city{Moscow}
  \country{Russia}
}
\email{roman.akinshin@skoltech.ru}

\author{Elizaveta Lopatina}
\orcid{0009-0005-2190-0846}
\affiliation{%
  \institution{Skolkovo Institute of Science and Technology}
  \city{Moscow}
  \country{Russia}
}
\email{elizaveta.lopatina@skoltech.ru}

\author{Kirill Bogatikov}
\orcid{0009-0005-4925-0777}
\affiliation{%
  \institution{Skolkovo Institute of Science and Technology}
  \city{Moscow}
  \country{Russia}
}
\email{kirill.bogatikov@skoltech.ru}

\author{Nikolai Kiz}
\orcid{0009-0000-4705-2703}
\affiliation{%
  \institution{Skolkovo Institute of Science and Technology}
  \city{Moscow}
  \country{Russia}
}
\email{nikolaj.kiz@skoltech.ru}

\author{Anna V. Makarova}
\orcid{0000-0002-6902-1630}
\affiliation{%
  \institution{MSU Institute for Artificial Intelligence}
  \city{Moscow}
  \country{Russia}
}
\email{makarovaav_03@my.msu.ru}

\author{Mikhail Lebedev}

\orcid{0000-0002-7457-837X}
\affiliation{%
  \institution{Lomonosov Moscow State University}
  \city{Moscow}
  \country{Russia}
}
\email{mikhail.lebedev@math.msu.ru}

\author{Miguel Altamirano Cabrera}
\orcid{0000-0002-5974-9257}
\affiliation{%
  \institution{Skolkovo Institute of Science and Technology}
  \city{Moscow}
  \country{Russia}
}
\email{m.altamirano@skoltech.ru}

\author{Dzmitry Tsetserukou}
\orcid{0000-0001-8055-5345}
\affiliation{%
  \institution{Skolkovo Institute of Science and Technology}
  \city{Moscow}
  \country{Russia}}
\email{d.tsetserukou@skoltech.ru}

\author{Valerii Kangler}
\orcid{0009-0004-4465-5372}
\affiliation{%
  \institution{Skolkovo Institute of Science and Technology}
  \city{Moscow}
  \country{Russia}}
\email{vkangler@motiv.ru}

\renewcommand{\shortauthors}{Akinshin et al.}

\begin{abstract}

  This paper presents a novel neuromorphic control architecture for upper-limb prostheses that combines surface electromyography (sEMG) with gaze-guided computer vision. The system uses a spiking neural network deployed on the neuromorphic processor AltAi to classify EMG patterns in real time while an eye-tracking headset and scene camera identify the object within the user’s focus. In our prototype, the same EMG recognition model that was originally developed for a conventional GPU is deployed as a spiking network on AltAi, achieving comparable accuracy while operating in a sub-watt power regime, which enables a lightweight, wearable implementation. For six distinct functional gestures recorded from upper-limb amputees, the system achieves robust recognition performance comparable to state-of-the-art myoelectric interfaces. When the vision pipeline restricts the decision space to three context-appropriate gestures for the currently viewed object, recognition accuracy increases to roughly 95\% while excluding unsafe, object-inappropriate grasps. These results indicate that the proposed neuromorphic, context-aware controller can provide energy-efficient and reliable prosthesis control and has the potential to improve safety and usability in everyday activities for people with upper-limb amputation.
\end{abstract}



\begin{teaserfigure}
\centering
  \includegraphics[width=0.9\textwidth, alt={The figure consists of two panels illustrating the gaze-guided EMG prosthetic control system. The left panel shows a first-person view from an eyeglass-mounted stereo camera overlooking a table with several objects. A red dot marks the user's current gaze point. The computer vision system identifies and labels objects (such as cups, bottles, or tools) and displays several recommended grasp types suited to the fixated object's shape and size—for example, a cylindrical grip for a cup or a pinch grip for a small item. The right panel depicts the next moment: the same first-person view now shows the user's prosthetic hand executing one of the recommended grasps as the user has confirmed their choice via muscle signals from an armband sensor. The hand is shown grasping the selected object, demonstrating the completed control loop from visual attention to motor execution.}]{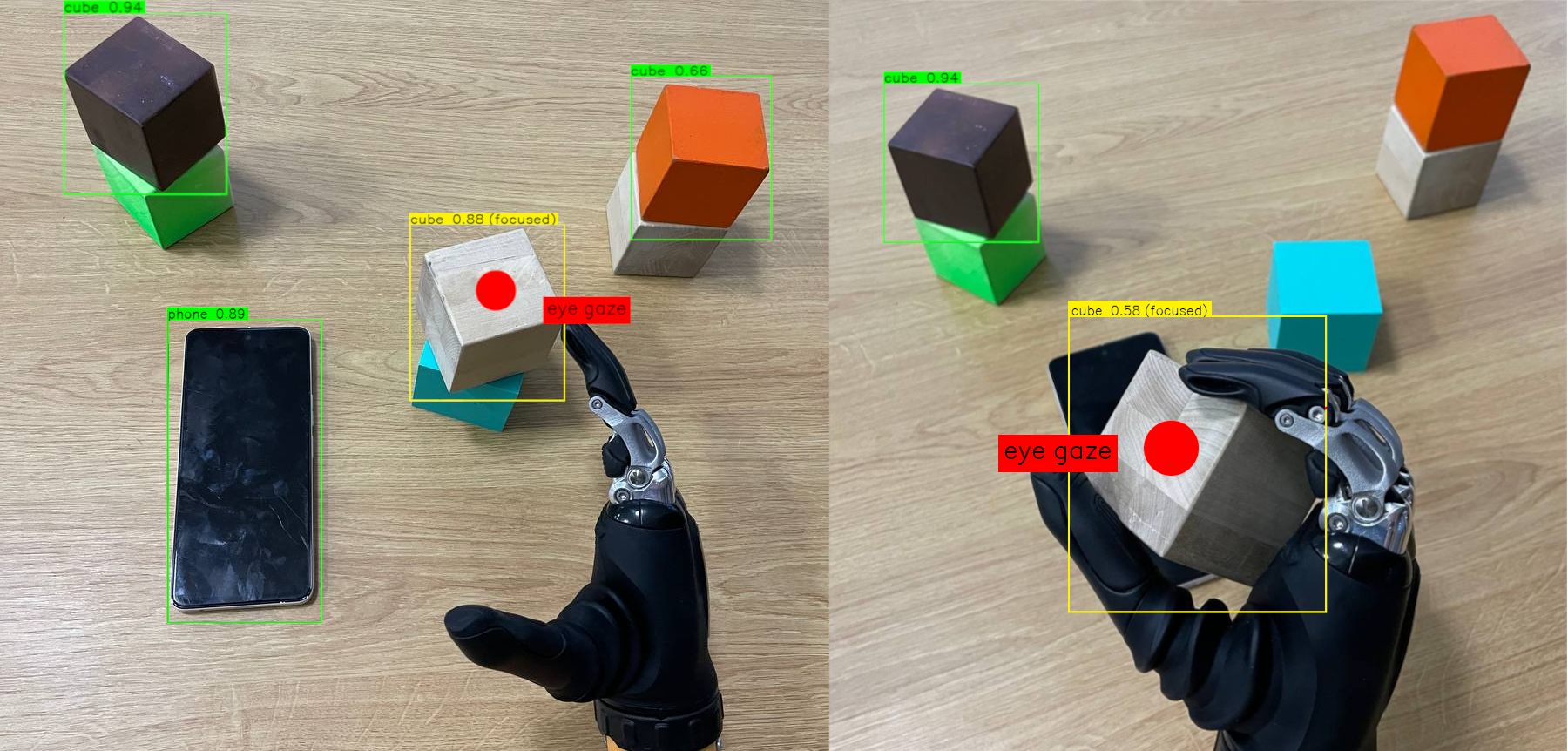}
  \caption{\textbf{Context-aware prosthetic control with EMG-based grasp selection and eye tracking}}
  \label{fig:teaser}
\end{teaserfigure}


\maketitle

\section{Introduction}

Controlling an upper-limb prosthesis is inherently safety-critical: a misclassified gesture or unintended activation can cause users to drop objects, damage their surroundings, or suffer physical injury. Despite advances in surface electromyography (sEMG), most prosthetic systems rely on a limited set of gestures due to signal variability and ambiguity. Moreover, many recent deep learning–based approaches achieve high classification accuracy at the cost of substantial energy requirements, often relying on GPUs that are difficult to integrate into lightweight, wearable devices \cite{li2021gesture}.

To address these limitations, neuromorphic computing has emer-ged as a promising paradigm for low-power, event-driven inference. Spiking neural networks (SNNs) deployed on neuromorphic hardware can achieve competitive performance on biosignal classification tasks while drastically reducing energy consumption \cite{pal2024npu}. This capability for continuous, low-latency decoding under strict power constraints makes neuromorphic EMG classifiers particularly well-suited for safety-critical prosthetic control, where always-on responsiveness is paramount.

At present, several complementary methods are available for gaze-based robot control\cite{Eeeesah,LLM-Glasses}.

In this work, we address these challenges by combining a neuromorphic EMG classifier with gaze-guided control system in a unified, context-aware control architecture. By leveraging event-driven neuromorphic inference for EMG decoding and using gaze-informed visual context to disambiguate user intent, our approach aims to improve operational safety while maintaining strict energy efficiency requirements. Compared to conventional GPU-based multimodal controllers, the proposed system is designed to be lightweight, scalable, and suitable for real-world deployment in wearable prosthetic devices.

\section{System Architecture}

The system consists of three main modules: (1) a multi-articulated hand prosthesis (MeHand by MaxBionic), (2) a surface electromyography (sEMG) armband mounted on the residual forearm, and (3) a head‑mounted vision module combining an eye tracker and stereo-cameras (Fig.2). (a) The prosthetic hand supports \textit{n} pre-programmed functional gestures. (b) The vision subsystem processes stereo RGB video and gaze position from the eyeglass-mounted eye tracker. Object detection via YOLO classifier identifies scene objects; the gaze intersection determines the currently fixated object and automatically predicts context-appropriate grasps, reducing the action space from \textit{n} to \textit{k} candidate gestures (\textit{k < n}). (c) The myoelectric armband decodes muscle signals to select the final gesture from the restricted subset of \textit{k} candidates. (d) User confirms selection via EMG activation, triggering prosthetic hand execution of the chosen grasp. The core idea is to use the visual context and current gaze position to infer a small task‑appropriate subset of hand gestures for a given object, then the sEMG interface selects one gesture from this reduced set and triggers the corresponding motion on the prosthesis \cite{zandigohar2024emg}.

A fully integrated vision system combining eye-tracking, synchronized camera inputs, and a YOLO object-recognition model continuously interprets the user’s gaze and the surrounding scene. It executes object detection and context‑to‑grasp inference and communicates high‑level gesture recommendations to the prosthesis controller \cite{cognolato2022multimodal}. A neuromorphic chip is applied to handle the demanding object-detection step with an event-driven YOLO-based neural network. A standard processor based on the Von Neumann architecture acts as a prefilter that redirects only selected areas of the image to the neuromorphic chip to match its limited input bandwidth. 

\subsection{Prosthetic Hand}

Current multi-articulated prosthetic hands offer high mechanical dexterity but suffer from a significant control bottleneck: standard myoelectric interfaces (typically 2-channel EMG) cannot intuitively map user intent to the device's high-dimensional actuation space (often more than 5 degrees of freedom). Users are frequently forced to cycle through preset grip patterns using unintuitive muscle co-contractions or mode switches, a cognitively demanding process that leads to high rejection rates \cite{xu2024poweredprosthetichandvision}.

Our system addresses this "dimensionality curse" by shifting grip selection to the context-aware architecture. We use the MeHand multifunctional myoelectric hand (MaxBionic), a multi-articulated end-effector with six actuators that enable independent digit flexion and thumb abduction (Fig. 2a). The device supports over 30 programmable grasp patterns (e.g., cylindrical, lateral, tripod) within a standard prosthetic form factor (approximately \SI{500}{g}, \SI{170}{mm} length) and provides a static load capacity of up to \SI{300}{N}\cite{maxbionic2025mehand}.

\begin{figure}[]
  \centering
  \includegraphics[width=1\linewidth, alt={The figure consists of 4 essential parts. First part describes initial gesture set for prosthetic hand. Second one describes computer vision system based on eye-tracking module that helps to perform object detection, which results in picking a smaller subset of gesture, compliant with classified object. Third part represents final gesture prediction based on EMG data. Last picture shows gesture execution.}]{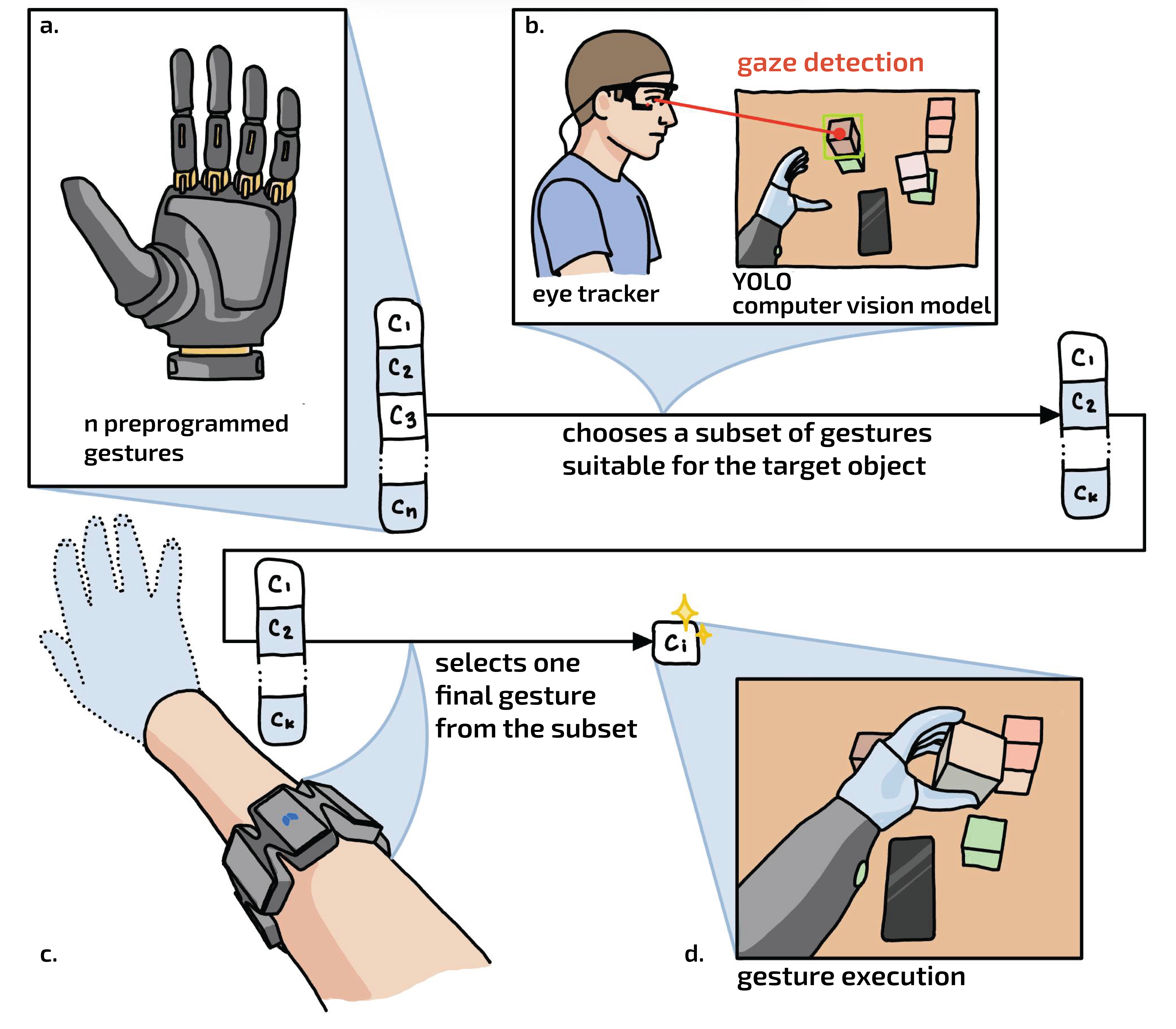}%
  \caption{\textbf{System architecture and control flow}}
  \label{fig:wide}
\end{figure}

\subsection{EMG}

The myoelectric interface is implemented using a Thalmic Labs Myo armband, which integrates eight dry EMG electrodes arranged circumferentially around the proximal forearm (Fig.3), providing multi‑channel surface EMG recordings (Fig.2c).

Each EMG channel is amplified and conditioned by an analog front-end with a band‑pass filter typically in the range of approximately from \SIrange{2}{40}{\hertz} to preserve the relevant myoelectric spectrum while attenuating motion artifacts and high‑frequency noise, combined with a narrowband notch filter at the local main frequency (50 Hz) to suppress power‑line interference \cite{quadrelli2025emg}. The signals are digitized at 200 Hz, a rate sufficient to capture voluntary muscle dynamics for pattern-recognition control.

\subsection{Computer Vision}
The computer vision subsystem comprises an eye tracker embedded in the eyeglass frame (using pupil-center–corneal-reflection to detect gaze direction and map it to a 3D gaze vector) and a forward-facing stereo RGB camera pair for object detection and depth estimation (Fig.2b). After intrinsic and extrinsic calibration, the gaze vector is transformed into the camera frame; the intersection of fixated objects' bounding boxes with the gaze ray identifies which object is currently fixated. The RGB input and computed depth map (via stereo matching and triangulation) are processed by a YOLO-based convolutional neural network to segment and classify scene objects. A secondary neural network maps the detected object class (e.g. cup, smartphone, door handle) and geometric attributes (size, orientation, reachable surface) to a small subset of candidate grasps from the prosthetic hand's gesture library, with associated probability scores. The EMG interface acts as a low-dimensional selector, either confirming the highest-probability grasp or cycling through alternatives under user control \cite{wang2022vision}.

To meet power and latency constraints on the wearable platform, the system is designed to offload the most demanding vision task — object detection — to a neuromorphic processor. A lightweight processor (e.g., Raspberry Pi) performs temporal differencing on the full-frame RGB input to detect regions with significant luminance change, constrained to a neighborhood around the gaze point, and extracts cropped Regions Of Interest (ROIs) that contain potential objects of interaction. Only these compact ROIs, rather than the full image, are encoded as event streams and transmitted to the neuromorphic chip,
where an SNN-style implementation of a YOLO - like model performs object detection over the incoming ROIs with low latency and power consumption while maintaining high detection accuracy for real-time hand–object interaction.

\section{System Evaluation}

\subsection{Neural network deployment on GPU and neuromorphic hardware}
Before designing the user study, we evaluated the computational characteristics of our EMG processing network on a conventional GPU and on the neuromorphic processor AltAi. The same trained artificial neural network was first executed on a desktop GPU (RTX 4080) and then converted to a spiking implementation for deployment on AltAi. For both platforms, we measured average power consumption and mean inference latency during continuous operation.

Prior work consistently reports that neuromorphic hardware consumes significantly less power than conventional GPU-based implementations for comparable tasks. Our measurements followed this trend: the GPU configuration required on the order of \SI{50}{W} with a latency of approximately \SI{5}{ms}, whereas the AltAi deployment operated at approximately \SI{0.07}{W} with a latency of roughly \SI{4.5}{ms}. This large reduction in power consumption motivated our subsequent user study, in which we examine how additional distal mass—a proxy for the heavier batteries and electronics required by higher-power systems—affects performance and fatigue in prosthesis users.

\begin{figure}[H]
    \centering
    \includegraphics[width=0.6\linewidth, alt={A male participant with forearm amputation sits on a chair wearing a myoelectric armband on his residual forearm. The armband is a sensor device with multiple electrodes that detect muscle signals for prosthetic control. The image shows the typical wearable form factor and placement of the EMG interface on the limb.}]{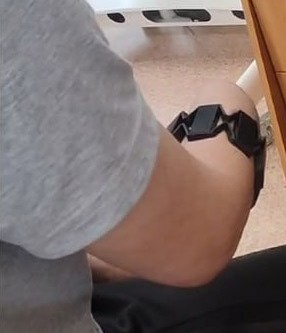}
    \caption{\textbf{Process of EMG data collection}}
    \label{fig:emg_setup}
\end{figure}

\subsection{Experimental Design}

We recruited 10 participants (6 male, 4 female) with forearm amputation and at least 6 months of daily prosthesis experience.The study protocol was approved by the Institutional Ethical committee (protocol №4, 30.09.2021). Informed consent was obtained from all participants. To evaluate the impact of electronics weight on user fatigue, we simulated three distal mass conditions: Light (\SI{100}{g}), Medium (\SI{200}{g}), and Heavy (\SI{300}{g}).

Participants performed a Discrete Transfer Task, moving 20 blocks (\SI{83}{g} each) from a table to a basket. Each condition consisted of three consecutive trials with rest intervals. We measured completion time and a Fatigue Index (the difference in time between the third and first trial). Subjective workload was assessed using the NASA-TLX questionnaire.

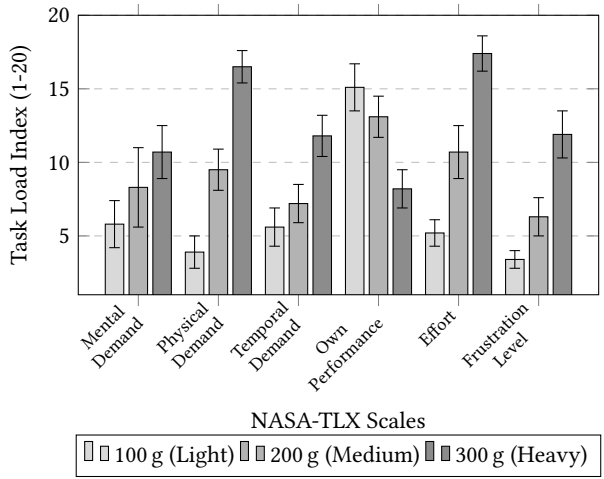
\begin{figure}[]
    \centering
    \Description{The figure represents bar chart with results of NASA-TLX questionnaire. A grouped bar chart displays NASA-TLX workload scores across six dimensions and three distal mass conditions. The horizontal axis lists six assessment scales: Mental Demand, Physical Demand, Temporal Demand, Own Performance, Effort, and Frustration Level. The vertical axis shows scores from 1 to 20. Three bar groups—representing 100 g (Light, light gray), 200 g (Medium, medium gray), and 300 g (Heavy, dark gray)—appear for each scale. Error bars indicate standard deviation ranges around each bar. Key observations: Mental Demand scores increase from approximately 5.8 (100 g) to 10.7 (300 g). Physical Demand shows the most pronounced increase, rising from 3.9 to 16.5 across conditions. Temporal Demand similarly increases from 5.6 to 11.8. Own Performance inversely decreases from 15.1 to 8.2, indicating reduced perceived performance with heavier loads. Effort escalates sharply from 5.2 to 17.4. Frustration Level increases from 3.4 to 11.9. Overall, the 300 g (Heavy) condition consistently shows elevated cognitive and physical demand alongside reduced perceived performance compared to lighter conditions.}
       
    \begin{tikzpicture}
        \begin{axis}[
            ybar,
            bar width=7pt,
            width=8.5cm,
            height=5.3cm,
            symbolic x coords={
                Mental,
                Physical,
                Temporal,
                Performance,
                Effort,
                Frustration
            },
            xtick=data,
            xticklabels={
                {Mental\\Demand},
                {Physical\\Demand},
                {Temporal\\Demand},
                {Own\\Performance},
                {Effort},
                {Frustration\\Level}
            },
            xticklabel style={
                rotate=45,
                anchor=east,
                align=center,
                font=\footnotesize
            },
            ylabel={Task Load Index (1-20)},
            xlabel={NASA-TLX Scales},
            ymin=1, ymax=20,
            enlarge x limits=0.15,
            legend style={
                at={(0.5,-0.5)},
                anchor=north,
                legend columns=-1
            },
            ymajorgrids=true,
            grid style=dashed,
        ]
            \addplot[
                fill=gray!30,
                error bars/.cd,
                y dir=both,
                y explicit
            ] coordinates {
                (Mental, 5.8) +- (0, 1.6)
                (Physical, 3.9) +- (0, 1.1)
                (Temporal, 5.6) +- (0, 1.3)
                (Performance, 15.1) +- (0, 1.6)
                (Effort, 5.2) +- (0, 0.9)
                (Frustration, 3.4) +- (0, 0.6)
            };

            \addplot[
                fill=gray!60,
                error bars/.cd,
                y dir=both,
                y explicit
            ] coordinates {
                (Mental, 8.3) +- (0, 2.7)
                (Physical, 9.5) +- (0, 1.4)
                (Temporal, 7.2) +- (0, 1.3)
                (Performance, 13.1) +- (0, 1.4)
                (Effort, 10.7) +- (0, 1.8)
                (Frustration, 6.3) +- (0, 1.3)
            };

            \addplot[
                fill=gray!90,
                error bars/.cd,
                y dir=both,
                y explicit
            ] coordinates {
                (Mental, 10.7) +- (0, 1.8)
                (Physical, 16.5) +- (0, 1.1)
                (Temporal, 11.8) +- (0, 1.4)
                (Performance, 8.2) +- (0, 1.3)
                (Effort, 17.4) +- (0, 1.2)
                (Frustration, 11.9) +- (0, 1.6)
            };

            \legend{\SI{100}{g} (Light), \SI{200}{g} (Medium), \SI{300}{g} (Heavy)}
        \end{axis}
    \end{tikzpicture}
    \caption{\textbf{Comparison of NASA-TLX scores across three weight conditions. Error bars represent standard deviation}}
    \label{fig:nasa_results}
\end{figure}

\subsection{Measures}
Upon completion of all three trials for each weight condition, participants completed the NASA Task Load Index (NASA-TLX) questionnaire. Ratings were based on their cumulative experience across the three trials of that weight condition. Specifically, the Physical Demand and Effort subscales were analyzed to evaluate the impact of added distal mass on perceived workload.

Throughout the experiment, some quantitative performance metrics were assessed, such as \textbf{completion time} - the time required to transfer all 20 blocks, measured in seconds - and \textbf{fatigue index} - the rate of change in completion time across trials, calculated as the difference between trial 3 and trial 1 within each weight condition. A positive value indicates fatigue accumulation.

\section{Results}

For six functional EMG gestures, our network achieved ~83\% accuracy in amputee participants with well-preserved residual muscles. Restricting the decision space to three context-appropriate gestures via gaze tracking increased accuracy to ~95\% and completely eliminated unsafe grasp attempts, demonstrating that context awareness directly enhances safety. In terms of efficiency, the neuromorphic processor AltAi implementation matched the GPU's low latency ($\approx4.5$ ms vs $\approx5$ ms) but operated at \SI{0.07}{W}, confirming its suitability for lightweight wearables.

The distal mass experiment quantified how additional prosthesis weight, originating from electronics and batteries, affects performance and perceived workload. In our data, task completion time increased for the heavier mass conditions, and the fatigue index was largest in the 300 g condition, indicating greater performance degradation across repetitions (Fig.5). NASA‑TLX ratings showed the same tendency: "Physical Demand" and "Effort" scores were higher on average for the \SI{300}{g} configuration than for the \SI{100}{g} baseline (Fig.4). These trends suggest that extra distal mass can worsen user performance and increase perceived effort, which supports the importance of energy-efficient controllers such as the AltAi processor for keeping prosthetic hardware lightweight.

\begin{figure}[H]
    \centering
    
    \includegraphics[width=1.0\linewidth, alt={The figure contains two line plots arranged as separate panels, the horizontal axis labeled “Distal Mass Condition” with three levels: 100 g, 200 g, and 300 g. On the left panel, the vertical axis shows Completion Time in seconds. Ten thin gray lines represent individual participants; each line connects three points, one for each mass level, corresponding to the mean time (over three repetitions) required to complete the block-transfer task. At 100 g, individual times range roughly from the mid‑40s to the high‑50s seconds. At 200 g, times increase into a band from the low‑60s to low‑80s seconds. At 300 g, times span approximately from 79 to 104 seconds. A thicker red line with square markers represents the group mean completion time: about 51.6 s at 100 g, 67.5 s at 200 g, and 92.1 s at 300 g. Overall, nearly all lines slope upward from left to right, showing that higher distal mass leads to slower task performance for almost every participant. On the right panel, the vertical axis shows Fatigue Index in seconds, defined as the difference in completion time between the third and first trial within each weight condition. Again, ten gray lines trace individual participants across the three mass levels. At 100 g, most Fatigue Index values lie between roughly 1 and 4 seconds, indicating little change across repetitions. At 200 g, values typically fall between about 3 and 6 seconds. At 300 g, indices are much larger, ranging from around 9 to 16 seconds, meaning that performance degrades more strongly over repeated trials when the hand is heavier. The red mean line shows an average Fatigue Index of approximately 2.5 s at 100 g, 4.7 s at 200 g, and 12.2 s at 300 g, highlighting a modest increase from 100 g to 200 g and a pronounced jump at 300 g. Together, the two panels demonstrate that added distal mass both slows overall performance and accelerates fatigue across repetitions.}]{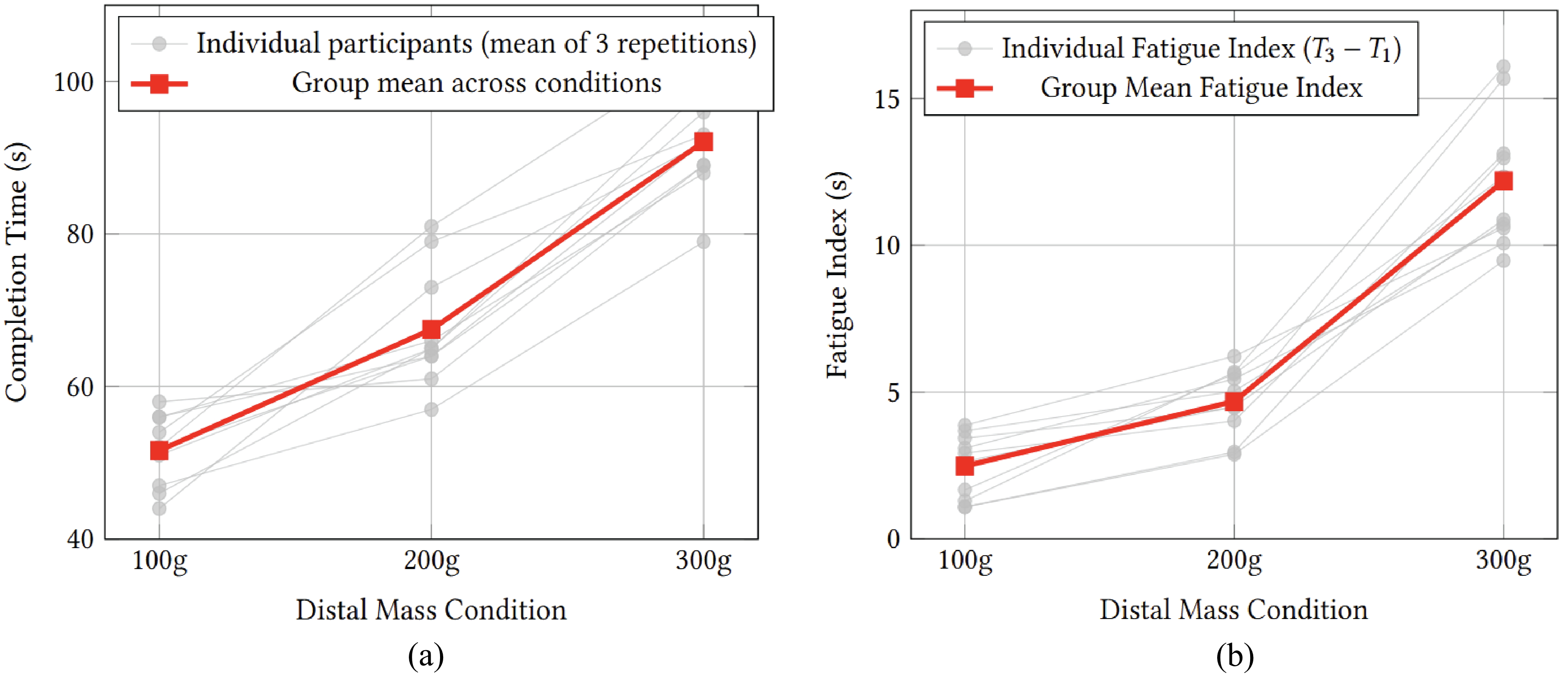}
    \caption{\textbf{Task performance and fatigue across distal mass conditions (a) Completion time for the block-transfer task. (b) Fatigue Index, defined as the change in completion time between the third and first trial within each condition}}

    \label{fig:emg_setup}
\end{figure}

\section{Conclusion and Future Work}
We presented a novel prosthesis control architecture that integrates neuromorphic EMG processing with gaze-guided context to enhance safety and wearability. To the best of our knowledge, this is the first system to implement this context-aware approach on a neuromorphic processor, achieving robust performance while operating in the sub-watt range. This extreme efficiency enables significant weight reduction, which our user study confirmed is critical for minimizing fatigue and maintaining performance.

Future work will focus on scaling this approach to more complex scenarios. We aim to migrate the full computer vision pipeline to the AltAi ecosystem to create a unified neuromorphic controller. We also plan to expand our energy benchmarks to include portable embedded GPUs, providing a more comprehensive baseline for wearable applications. Additionally, we will leverage next-generation processors to implement on-chip online learning, enabling real-time adaptation without offline retraining. Finally, we will rigorously validate these adaptive algorithms on a larger cohort with diverse amputation levels to ensure robustness across variable physiological conditions.

\section*{Acknowledgments}

This work was supported by the MSU Institute for Artificial Intelligence.

\balance

\end{document}